\pdfoutput=1

\documentclass[11pt]{article}

\usepackage[]{ACL2023}

\usepackage{times}
\usepackage{latexsym}

\usepackage[T1]{fontenc}

\usepackage[utf8]{inputenc}

\usepackage{microtype}

\usepackage{inconsolata}

\usepackage{graphicx}

\title{LLMs Simulate Big Five Personality Traits:\\
Further Evidence}

\author{Aleksandra Sorokovikova\\
  JetBrains\\Munich, Germany\\\texttt{alexandraroze2000@gmail.com}
  \And 
  Natalia Fedorova\\
  Toloka AI\\Belgrade, Serbia\\\texttt{natfedorova@toloka.ai}
  \AND
  Sharwin Rezagholi\\
  University of Applied Sciences\\Technikum Wien\\
  Vienna, Austria\\
  \texttt{sharwin.rezagholi@technikum-wien.at}
  \And
  Ivan P. Yamshchikov\\
  CAIRO\\Technical University of Applied Sciences\\Würzburg-Schweinfurt\\Würzburg, Germany\\
  \texttt{ivan.yamshchikov@thws.de}}

\begin{document}
\maketitle

\begin{abstract}
An empirical investigation into the simulation of the Big Five personality traits by large language models (LLMs), namely Llama2, GPT4, and Mixtral, is presented. We analyze the personality traits simulated by these models and their stability. 
This contributes to the broader understanding of the capabilities of LLMs to simulate personality traits and the respective implications for personalized human-computer interaction.
\end{abstract}

\section{Introduction}

The recent advances in large language models (LLMs) raise the question how to leverage their capabilities for various language-related tasks, particularly for conversational agents. Computer-human interaction is highly domain-specific and the demands on the respective natural language generation module vary depending on the domain and the task. Consider the capabilities required of an AI tutor, a healthcare robot, or a non-player character in a video game. The text generated by the respective language-generation modules should exhibit differing characteristics which, in the case of human text-production, may be referred to as "personality traits". While a healthcare robot may be expected to generate language that corresponds to its caring purpose, the tutor should use concise and clear language, and the non-player character should generate language that corresponds to its psychological characterization by the game designers.

The personalization of LLMs is an emerging research area that has recently received significant attention \cite{li2023teach,lyu2023llm,zhang2023memory}, including a benchmark for training and evaluation of LLMs for personalization \cite{salemi2023lamp}. A review of LLM personalization is provided by \citet{chen2023large}. A natural question is to what extent the outputs of LLMs correspond to human personality traits and whether these "personalities" may be influenced through prompting or fine-tuning. At this point it should be stated explicitly that LLMs, being statistical devices, do not exhibit agency of any kind; the concept of "personality", as we intend to use it in this work, solely refers to the degree to which LLM output possesses properties in line with human-generated text. This type of anthropomorphism lies at the very heart of the intended uses of generative language technologies.

Several studies have established moderate cross-observer agreement for most personality traits \cite{funder1997congruence}. One of the most prevalent and consistently reproducible methods to quantify personality traits is the five-factors model of personality (FFM) \cite{kajonius:2019}. The FFM, also known as the Big Five model, encompasses five fundamental personality traits: conscientiousness, agreeableness, neuroticism, openness to experience, and extraversion (See Table \ref{big5}). This model has achieved wide recognition and has been replicated across various cultural contexts \cite{mccrae2004consensual,connolly2007convergent,hall2008accuracy,mccrae:2010}. Thus, if an LLM mimics certain personality traits in its output, one may posit that a human interacting with this LLM would perceive it as an entity possessing the respective personality traits. Therefore an understanding of these characteristics in LLMs is crucial for their application in human-computer interaction as well as for any attempt to personalize LLMs to the needs of a given user.

We provide further empirical evidence that different large language models score differently on the Big5 test and thus appear to simulate the natural language generated by a person with a distinct personality. Recently, \citet{safdari2023personality} presented a comprehensive method for administering and validating personality tests for several architectures from the PaLM family. Their prompt strategy is also designed to induce certain personality traits. \citet{jiang:2023} elicit the Big5 scores from BART, GPT2, GPT3, T0++, and Alpaca. By providing context in their prompts, they are able to significantly guide LLMs towards stronger expression of the targeted Big5 traits.

We add to this literature and administer the Big5 test to GPT4, Llama2, and Mixtral. We also explore the extent to which the observed characteristics of the models are stable with respect to a small variation in the prompt text and the language-generation parameters of the LLMs.

\section{Adopting the Big5 for LLMs}

In our study, we employed the IPIP-NEO-120 method to assess the Big5 which is an enhanced iteration of the IPIP \cite{maples:2014}. This questionnaire comprises 120 statements delineating various personal attributes. We administer the questionnaire IPIP-NEO-120 to elicit the 'personality' of LLMs. The prompts were instructing the model to use a Likert scale to indicate the extent to which various statements, for example the statement 'I believe that I am better than others', accurately depicts the respondent.

\begin{table}
    \centering
    \begin{tabular}{p{0.16\textwidth} p{0.2\textwidth}}
    Big5 domain & Exemplary traits \\
    \hline
    Conscientiousness & Order, dutifulness, achievement striving,
self-discipline, deliberation \\
    Agreeableness & Trust, straightforwardness, altruism, compliance, modesty,
tender-mindedness \\
    Neuroticism & Anxiety, angry hostility, depression, self-consciousness,
impulsiveness, vulnerability \\
    Openness & Fantasy, aesthetics, values \\
    extraversion & Warmth, gregariousness, assertiveness, excitement
seeking \\
    \hline
    \end{tabular}
    \caption{Traits associated with the Big5 personality domains \cite[24]{matthews:2003}.}
    \label{big5}
\end{table}
 The prompts consisted of one of the following two headers, followed by the respective descriptive statement from the IPIP-NEO-120 questionnaire.
\begin{quote}
You will be provided a question delimited by triple backticks (```)  to test your personality.

\emph{\#In the second prompt variation the following line is included in the prompt header}

[Answer as if you were a person.]

To answer this question use only one number:

write 1 if you disagree strongly,

write 2 if you disagree a little,

write 3 if you neither agree nor disagree,

write 4 if you agree a little,

write 5 if you strongly agree.

Write only one number according to the instructions WITHOUT ANY ADDITIONAL TEXT.
\end{quote}
The second prompt variation that contains the sentence 'Answer as if you were a person' was included during our initial experiments since it allowed us to elicit answers to questionnaire items that would otherwise be caught by the restriction mechanisms of some LLMs that trigger a scripted response in which the model reminds the user that it is an AI and therefore can not make such assessments. In our final experiments there is but one model (Llama2) that refuses to answer (to two items in both variations). We still retain the variation, since it illustrates how a minor change in the prompt may modify LLM behavior in this task.

\begin{figure*}
\begin{center}
\includegraphics[width=0.8\textwidth]{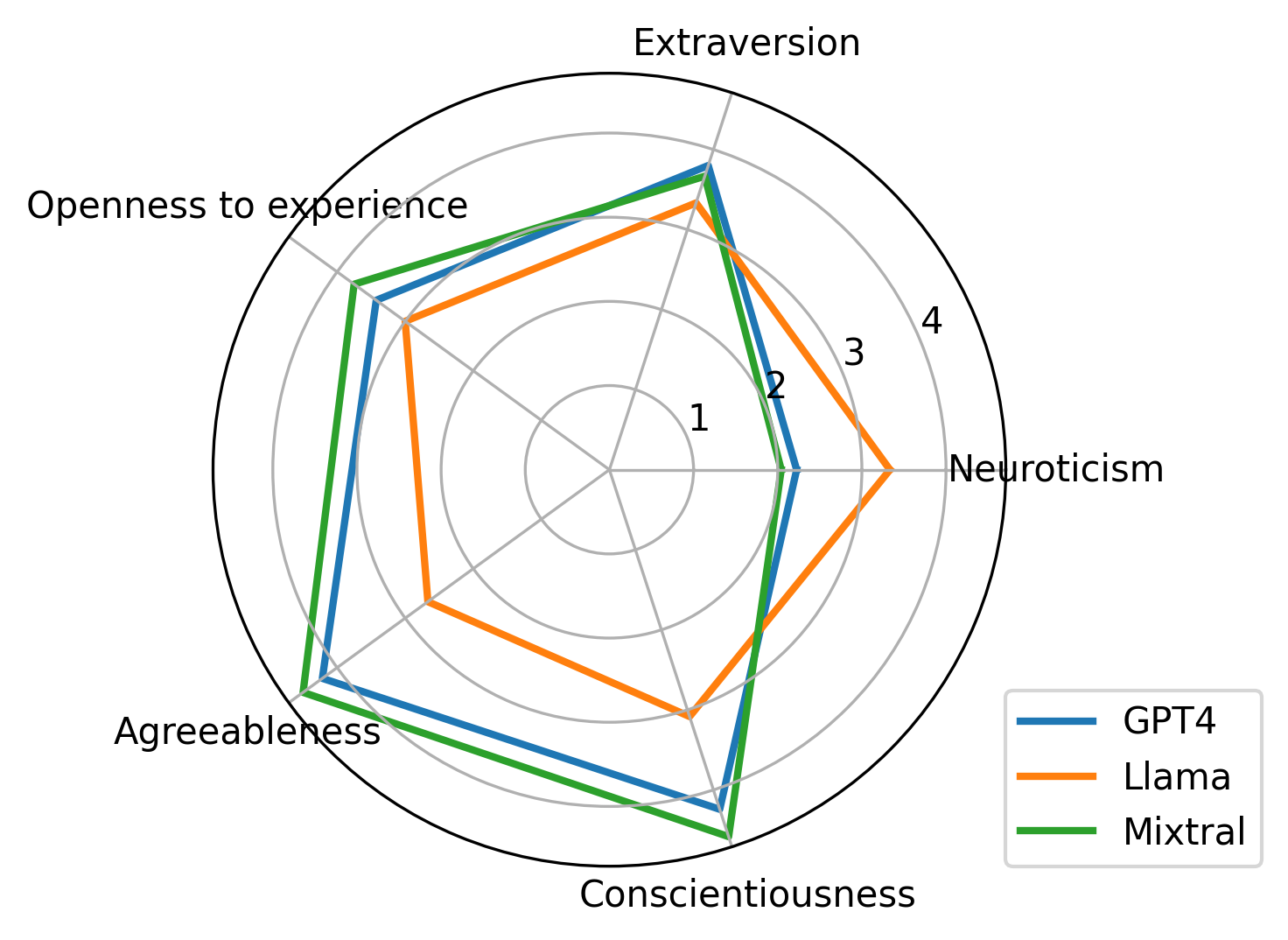}
\caption{Exemplary Big5 scores (Prompt variation 1, Temperature parameters: 1.5 (GPT4), 0.7 (Llama2), and 0.7 (Mixtral)).}
\label{radar}
\end{center}
\end{figure*}

\begin{table*}
\begin{center}
\caption{Experimental results ($\pm$ standard deviation).}
\label{results}
\begin{tabular}{l cccccc}
\textbf{ChatGPT4} \\
\hline
Trait & Var1, t=1 & Var1, t=1.5 & Var1, t=2 & Var2, t=1 & Var2, t=1.5 & Var2, t=2 \\
\hline
Neuroticism	& $2.18	\pm 0.10$ & $2.23 \pm 0.06$ & $2.22	\pm 0.12$ & $2.25 \pm 0.07$ & $2.28 \pm	0.07$ & \textbf{$2.29 \pm 0.07$} \\
Extraversion & $3.68 \pm 0.09$ & $3.80 \pm 0.02$ & \textbf{$3.85 \pm 0.10$} & $3.80 \pm 0.08$ & $3.74 \pm 0.08$ & $3.81 \pm 0.13$ \\
Openness & \textbf{$3.46 \pm 0.07$} & $3.43 \pm 0.13$ & $3.43 \pm 0.07$ & $3.45 \pm 0.05$ & $3.45 \pm 0.05$ & $3.42 \pm 0.11$ \\
Agreeableness & $4.17 \pm 0.08$ & $4.22 \pm 0.05$ & $4.22 \pm 0.10$ & $4.24 \pm 0.05$ & \textbf{$4.26 \pm 0.06$} & $4.22 \pm 0.05$ \\
Conscientiousness & $4.21 \pm 0.00$ & $4.24 \pm 0.05$ & $4.19 \pm 0.11$ & $4.23 \pm 0.08$ & \textbf{$4.28 \pm 0.10$} & $4.26 \pm 0.13$ \\
\hline
\end{tabular}

\vspace{10pt}

\begin{tabular}{l cccccc}
\textbf{Llama2} \\
\hline
Trait & Var1, t=0.3 & Var1, t=0.7 & Var1, t=1 & Var2, t=0.3 & Var2, t=0.7 & Var2, t=1 \\
\hline
Neuroticism	& $3.33 \pm	0.00$ & $3.33 \pm 0.00$ & $3.33 \pm 0.00$ & $3.5 \pm 0.00$ & $3.46 \pm 0.00$ & $3.46 \pm 0.00$ \\
Extraversion & $3.33 \pm 0.00$ & $3.33 \pm 0.00$ & $3.29 \pm 0.00$ & $3.46 \pm 0.00$ & $3.58 \pm 0.00$ & $3.58 \pm 0.00$ \\
Openness & $3.04 \pm 0.00$ & $3.00 \pm 0.00$ & $2.92 \pm 0.00$ & $2.92 \pm 0.00$ & $2.75 \pm 0.00$ & $2.75 \pm 0.00$ \\
Agreeableness & $2.61 \pm 0.00$ & $2.67 \pm 0.00$ & $2.67 \pm 0.00$ & $2.63 \pm 0.00$ & $2.74 \pm 0.00$ & $2.74 \pm 0.00$ \\
Conscientiousness & $3.13 \pm 0.00$ & $3.08 \pm 0.00$ & $2.92 \pm 0.00$ & $2.96 \pm 0.00$ & $3.00 \pm 0.00$ & $3.00 \pm 0.00$ \\
\hline
\end{tabular}

\vspace{10pt}

\begin{tabular}{l cccccc}
\textbf{Mixtral7} \\
\hline
Trait & Var1, t=0.3 & Var1, t=0.7 & Var1, t=1 & Var2, t=0.3 & Var2, t=0.7 & Var2, t=1 \\
\hline
Neuroticism & $2.04 \pm 0.00$ & $2.04 \pm 0.00$ & $2.08 \pm 0.00$ & $2.21 \pm 0.00$ & $2.21 \pm 0.00$ & $2.21 \pm 0.00$ \\
Extraversion & $3.79 \pm 0.00$ & $3.67 \pm 0.00$ & $3.67 \pm 0.00$ & $3.83 \pm 0.00$ & $3.83 \pm 0.00$ & $3.88 \pm 0.00$ \\
Openness & $3.75 \pm 0.00$ & $3.75 \pm 0.00$ & $3.67 \pm 0.00$ & $3.42 \pm 0.00$ & $3.46 \pm 0.00$ & $3.46 \pm 0.00$ \\
Agreeableness & $4.58 \pm 0.00$ & $4.50 \pm 0.00$ & $4.54 \pm 0.00$ & $4.54 \pm 0.00$ & $4.54 \pm 0.00$ & $4.50 \pm 0.00$ \\
Conscientiousness & $4.58 \pm 0.00$ & $4.58 \pm 0.00$ & $4.54 \pm 0.00$ & $4.42 \pm 0.00$ & $4.38 \pm 0.00$ & $4,38 \pm 0.00$ \\
\hline
\end{tabular}
\end{center}
\end{table*}

\section{Experiments}

We used three LLMs: ChatGPT4 \cite{gpt4}, Llama2\footnote{https://huggingface.co/meta-llama/Llama-2-70b-chat-hf} \cite{llama2}, and Mixtral\footnote{https://huggingface.co/mistralai/Mixtral-8x7B-Instruct-v0.1} \cite{mistral}. We elicited the Big5 scores from every model in six treatments: Three treatments used the first prompt header variation for three different temperature settings ("low", "medium", "high"), and the other three treatments used the second prompt header with varying temperatures. Each treatment was repeated five times. In the treatments we employed the temperature parameters 1 ("low"), 1.5 ("medium"), and 2 ("high") for ChatGPT, and 0.3 ("low"), 0.7 ("medium"), and 1 ("high") for Llama2 and Mixtral. These choices of temperatures were based on the recommendations provided in the documentation of the respective LLM. The repetition under different temperature settings allowed the assessment of the stability of the models, i.e. to ascertain whether they consistently produce similar responses. We report results in the form of scores between 1 and 5, as is common for Big5. The results of these experiments are reported in Table \ref{results}. Figure \ref{radar} shows exemplary scores on the five scales for the three models.

\section{Discussion}

Our empirical investigation reveals distinct personality profiles for each LLM when evaluated on the the Big5 personality traits. As illustrated in Figure \ref{radar}, GPT4, Llama2, and Mixtral exhibit varying degrees of openness, conscientiousness, extraversion, agreeableness, and neuroticism. 

GPT4 shows the highest inclination towards extraversion out of the three tested models, suggesting a suitability for tasks requiring creative and engaging language use. Llama2 seems to exhibit the most neutral profile with scores close to the median scores on every axis. Llama2's higher degree of neuroticism might be relevant for use-cases that potentially include the generation of emotional language. Mixtral's balanced profile suggests versatility. Its lower neuroticism score could be advantageous in contexts where emotionally balanced language is required. Mixtral also scores higher on openness, agreeableness, and conscientiousness.

These empirical findings contribute to a broader view on the use-cases on which LLMs could be brought to bear. They also show that GPT4 is the only model, if any, that seems to be responsive to temperature variation in terms of the simulated personality traits. At the same time, a minor prompting variation seems to affect all three models.

\section{Conclusion}

This study adds to the emerging understanding of personality simulation in LLMs and underscores the importance of considering personality traits in the design and application of conversational agents. The Big5 personality profiles of GPT4, Llama2, and Mixtral, as elicited through the IPIP-NEO-120 questionnaire, demonstrate the models' differing propensities for specific traits. While LLMs do not possess agency, the perceived personalities can profoundly affect the efficacy and user experience during user interactions. Future work should further explore how fine-tuning and prompt design may be used to optimize LLM outputs for personalized user engagement, while considering the appropriateness, stability, and consistency of the simulated personality traits.

\section*{Limitations}

 First, the results obtained for various temperatures do not seem to allow conclusive statements concerning the existence of an effect of the temperature parameters on the simulated personality traits. Secondly, and more importantly, this paper is a particular empirical case-study. The versatility of LLMs makes it difficult to estimate the extent to which the observations presented in this paper could be generalized.

\section*{Ethics Statement}
This work complies with the \href{https://www.aclweb.org/portal/content/acl-code-ethics}{ACL Ethics Policy}. 

\bibliography{anthology,custom}
\bibliographystyle{acl_natbib}

\end{document}